\documentclass[letterpaper]{article}
\usepackage{iccc}
\usepackage{tabu}
\usepackage{times}
\usepackage{helvet}
\usepackage{courier}
\usepackage{graphicx}
\title{Creative Invention Benchmark}
\author{Matthew Guzdial, Nicholas Liao, Vishwa Shah, and Mark O. Riedl\\
School of Interactive Computing\\
Georgia Institute of Technology\\
Atlanta, GA 30332  USA\\
mguzdial3@gatech.edu, nliao7@gatech.edu, vishwashah@gatech.edu, riedl@cc.gatech.edu\\
}
\begin{document} 
\maketitle
\begin{abstract}
\begin{quote}
In this paper we present the Creative Invention Benchmark (CrIB), a 2000-problem benchmark for evaluating a particular facet of computational creativity. Specifically, we address combinational p-creativity, the creativity at play when someone combines existing knowledge to achieve a solution novel to that individual. We present generation strategies for the five problem categories of the benchmark and a set of initial baselines. 
\end{quote}
\end{abstract}

\section{Introduction}

Benchmarks represent a common means for driving community effort on a specific task. For example, MNIST is a dataset of handwritten digits paired with their numeric value, which proved popular and produced breakthroughs in the field of image recognition \cite{lecun1998mnist}. At present it is considered solved by modern methods, but has continued as a standard for evaluating novel approaches, given that there are known performances for comparable techniques. While imperfect, we posit a benchmark for creativity could accomplish similar effects for the field of computational creativity. Creativity itself is too ill-defined for any benchmark. But just as recognizing handwritten digits does not translate to mastery of computer vision, we can define a set of creative tasks to address one particular facet of creativity.

Imagine you are an agent with a limited knowledge base. You know about the color red (255,0,0) and blue (0,0,255), and you know that there are integer variables $x$ and $y$ that can range from 0 to 100. You have access to a single method \textbf{Paint}, that given a value for $x$ and $y$, and a color (represented in RGB) paints a pixel of a canvas. In return for painting a pixel the agent receives a floating point score [0-1] that grades the agents current painting compared to an unseen goal. The goal, in this case, is to paint a picture of a grape.

As a human reading the problem description above, the answer to this problem appears obvious. From red and blue make purple, or perhaps multiple shades of purple, in order to paint the unseen picture of a grape. This instinct can be understood as an instantiation of what Boden calls combinational creativity \cite{boden2004creative}. But it does not reflect how a naive AI agent might solve this problem, instead greedily placing blue and red to maximize the score to some local maximum without inventing the color purple. Alternatively one might naively hand-author all possible colors for the AI agent, but this would run against the spirit of the problem. Solving this problem clearly requires creativity, but we do not argue that this problem can evaluate the entirety of creativity. We instead focus on {\em p-creative}, {\em combinational} creativity \cite{boden1998creativity}. P-creativity refers to creation of artifacts that are novel to the individual creator based on its knowledge (e.g., the artifact could have been invented by other creators previously). Combinational creativity refers to the creation of artifacts through the process of recombining existing knowledge. For the purposes of this paper we refer to this class of problem as invention problems.

In this paper we present the Creative Invention Benchmark (CrIB)\footnote{https://github.com/mguzdial3/CrIB}, a publicly available benchmark of 2000 problems in 5 domains (painting, alien language, photobashing, narrative, and dessert recipes). All of these problems fit the general form of the painting example, requiring an agent to generalize and invent new concepts from a given problem-specific knowledge base to reach a solution given feedback from an unseen goal. 

The example of painting an unseen grape may seem trivial but it is analogous to many of the most interesting and practical problems currently facing society from product invention to drug discovery. As humans we reflect on our existing knowledge to invent radical solutions, and we anticipate a need for artificial agents to do the same. 

An important---but largely overlooked---challenge in computational creativity is cross-domain creativity, wherein a single agent or model is able to address creative problems from disparate domains. 
Commonly creativity researchers sidestep the need for general creative reasoning through hand-authoring of domain-specific knowledge.
To the best of our knowledge this represents the first such cross-domain benchmark for computational creativity.

The rest of this paper is organized as follows: In section two we discuss related work and historic work that informs our position for this paper. In section three we discuss CrIB, all five problem categories and examples of each problem. In section four we demonstrate various baselines and features of the benchmark. We end with a discussion of the limitations of the benchmark, applications, and future directions.

\section{Related Work}

\subsection{Creativity Tests}

There exist prior formal tests that involve computational models of creativity. For example the Lovelace 1.0 \cite{bringsjord2003creativity}, Lovelace 2.0 \cite{riedl2014lovelace} and MacGyver tests \cite{sarathy2017macgyver} formalize bounds and loose evaluations that require creative cognition. However none of these prior approaches present sets of individual problems. Ravens Progressive Matrices \cite{raven1938raven} has been used as a test for general cognitive ability, which includes creativity, most notably in work such as \cite{shegheva2018}. However, this test does not specifically seek to test creativity and only makes use of a single domain, whereas CrIB focuses on cross-domain, combinational p-creativity.

\subsection{Combinational Creativity}

There exists a range of combinational creativity techniques, which we briefly summarize. Notably researchers of combinational creativity do not frequently self-identify as addressing the same problem or field. Thus many combinational creativity approaches remain dependent on particular problem domains. However there has been some recent work to attempt to tie this field together \cite{guzdialcombinatorial}.

Case-based reasoning (CBR) represents a general AI problem solving approach that relies on the storage, retrieval, and adaption of existing solutions \cite{de2005retrieval}. The adaption function has lead to a large class of combinational creativity approaches, falling in two categories of either substitutional or structural adaption \cite{wilke1998techniques,fox2009exploring}. These techniques tend to be domain-dependent, for example for the problem of text generation or tool creation \cite{hervas2006case,sizov2015evidence}.

Genetic Algorithms (GAs) represents a general AI problem solving approach that relies on an abstracted model of biological evolution \cite{srinivas1994genetic}. It has proven extremely popular among computational creativity practioners, and we make use of it for an initial agent for solving CrIB. While not often recognized as such, the crossover function of a GA can be understood as a combinational creativity approach \cite{herrera2003taxonomy}, though as with CBR adaption crossover functions tend to be domain-dependent.

Beyond CBR and GAs the area of belief revision, modeling how beliefs change, includes a function to merge existing beliefs with new beliefs\cite{konieczny2004da2,steels2006unify,cojan2008conservative,cojan2009belief,konieczny2011logic}. The mathematical notion of convolution has also been applied to blend weights, but with inconclusive results \cite{thagard2011aha}.
 
 \begin{figure*}[tb]
  \centering
  \includegraphics[height=1.7in]{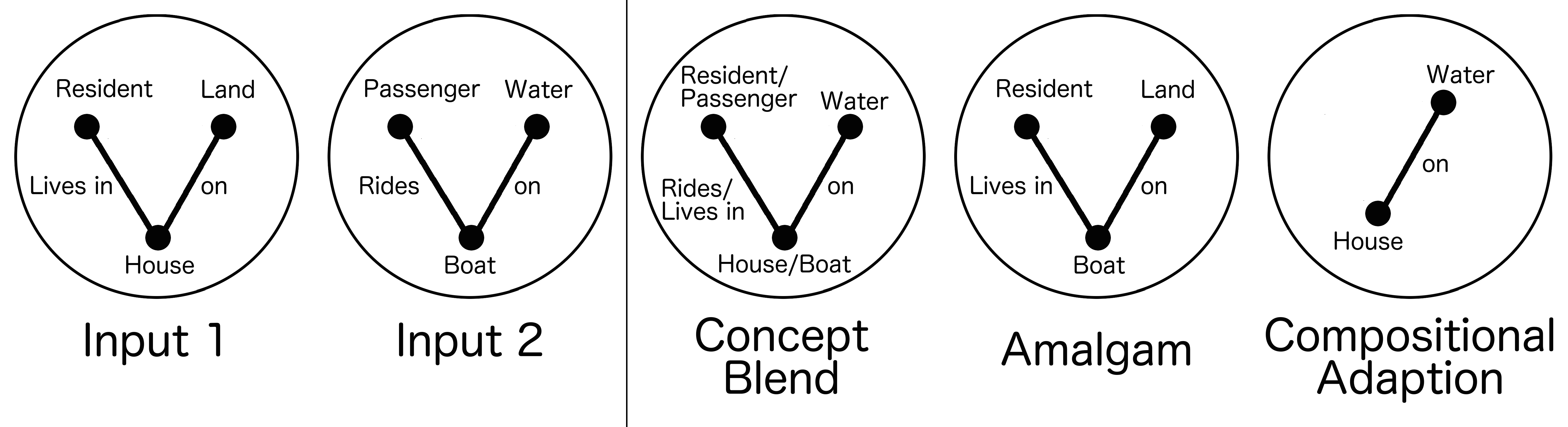}
  \caption{Example of three combinational creativity techniques. Two input spaces on left with example output from the three techniques on the right.}
  \label{fig:combinatorialExamples}
\end{figure*}
 
We identify three combinational creativity approaches for further discussion given their popularity and generality across multiple problem domains. We visualize these approaches with illustrative examples in Figure \ref{fig:combinatorialExamples}.

\subsubsection{Concept Blending}

Fauconnier and Turner \shortcite{fauconnier1998conceptual} formalized the ``four space" theory of concept blending. They described four spaces: two \textit{input spaces} represent the unblended elements, input space points are projected into a common \textit{generic space} to identify equivalence, and these equivalent points are projected into a \textit{blend space}. In the blend space, novel structure and patterns arise from the projection of equivalent points. Fauconnier and Turner \cite{fauconnier1998conceptual,fauconnier2008way} argued this was a ubiquitous process, occurring in discourse, problem solving, and general meaning making. 

Concept blending typically requires a large amount of human authoring for individual concept spaces. More recent work has looked into automatically learning or deriving concepts \cite{o2015stimulating,guzdial-iccc2016}. There has been work in blending individual tagged exemplars together based on surface level features of components \cite{alhashim2014topology}. Fauconnier and Turner originally developed a set of heuristics for domain-independent measures of quality for blends, while more recent work has looked to introduce goals for blends \cite{li2012goal}.

\subsubsection{Amalgamation}

Onta{\~n}{\'o}n and Plaza designed amalgams as a formal unification function between multiple cases \cite{ontanon2010amalgams}. Similar to concept blending, amalgamation requires a knowledge base that specifies when two components of a case share a general form, for example ``French" and ``German" both share the more general form ``nationality". Unlike concept blending, this shared generalization does not lead to a merging of components, but requires that only one of components be present in a final amalgam. For example, a ``red French car" and an ``old German car" could lead to an ``old red French car" or an ``old red German car". 

Amalgams have been utilized as the adaption function in CBR systems \cite{manzano2011amalgam}, combined with concept blending for product development \cite{besold2015generalize}, and adapted to an asymmetrical form for story generation \cite{ontanon2012case}. Amalgamation represents a strong general method for combinational creativity. However it suffers from the drawbacks of other methods in terms of a traditional reliance on authored knowledge bases and domain-specific generalization.

\subsubsection{Compositional Adaption}

Compositional adaption arose as a CBR adaption approach \cite{holland1989induction,fox2009exploring}, but has found significant applications in adaptive software \cite{mckinley2004taxonomy,eisenbach2007component}. The intuition behind compositional adaption is that individual concept components can be broken apart and recombined based on their connections. In adaptive software this process takes sets of functions with given inputs and outputs, and strings them together to achieve various effects, which makes compositional adaption similar to planning given a goal state or output. However, it can also be applied in a goal-less way to generate valid compositions. 

Compositional adaption has been applied to recipe generation \cite{muller2014compositional,badie2017compositional}, intelligent tutoring systems \cite{reyhani2003new}, and traditional CBR approaches \cite{chedrawy2006case}. Unlike other methods compositional adaption does not require an explicit generalization knowledge base. However, it is common to make use of a knowledge base to generalize across components and their relationships in order to expand the set of valid combinations.

\section{Creative Invention Benchmark (CrIB)}

In this section we discuss in more detail the Creative Invention Benchmark (CrIB). Our goal for the benchmark is to evaluate goal-driven combinational p-creativity, meaning a creative problem solving technique that relies on recombining the knowledge available to an individual. We refer to this class of problems as invention problems. To address our goal of generality we test combinational p-creativity across five distinct domains. This further reflects the multidisciplinary field of computational creativity. The domains are:
\begin{enumerate}
  \item {\bf Painting}, as in the running example in the introduction, in which an agent must invent new colors from some initial knowledge base to approximate some unknown goal painting.
  \item {\bf Alien language}, in which an agent must invent novel words to recreate an unknown goal sentence.
  \item {\bf Photobashing}, a practice from the field of concept art in which existing images are pieced together to form novel art. In this problem domain the agent must combine input images to approximate some unknown goal photobash.
  \item {\bf Narrative}, in which an agent, given a graphical representation of at least two story domains, must tell a target unknown goal story in some novel domain.
  \item {\bf Dessert Recipes}, in which an agent must combine existing recipe ingredients to create an unknown goal recipe.
\end{enumerate}

The benchmark has a total of 2000 problems evenly spread across the five domains for a total of 400 problems per domain. For each problem an agent receives an initial knowledge base, a function to apply the agent's knowledge base in a domain-appropriate way (e.g. adding words to a sentence in the alien language domain), a function to clear the current state of the agent's submission in a domain-appropriate way (e.g. resetting the current canvas to a blank canvas in the painting domain), and a scoring function that measures the agent's distance to some unknown goal (with values ranging from 0.0 to 1.0). 

In the following subsections we discuss each domain in further detail. Notably we discuss the structure of each problem in terms of the input knowledge base, functions available to the agent, and the problem goal. We also discuss the approach taken to generate the domain problems and demonstrate an example problem. We note that all relevant code can be found at the public CrIB GitHub.

\subsection{Painting}

\begin{figure}[tb]
  \centering
  \includegraphics[width=2.5in]{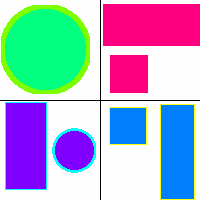}
  \caption{Four examples of unseen goal ``paintings''}
  \label{fig:paintingTargets}
\end{figure}

\begin{figure*}[tb]
  \centering
  \includegraphics[height=1.5in]{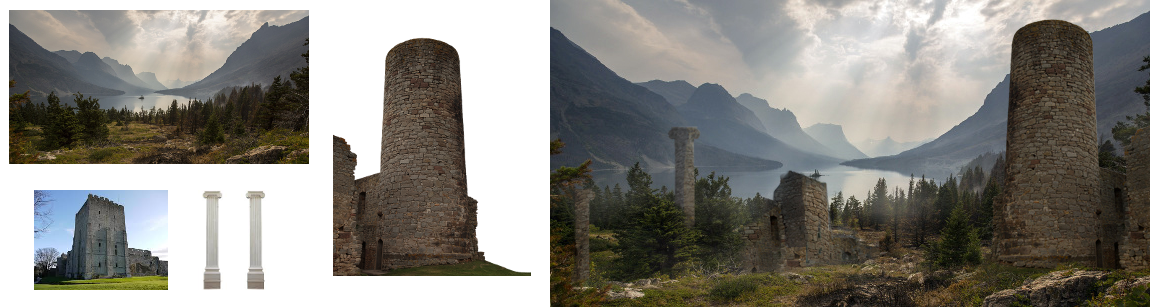}
  \caption{Example of a human photobash from one of our artists on the right with the input images used on the left.}
  \label{fig:humanPhotobash}
\end{figure*}
\begin{figure}[tb]
  \centering
  \includegraphics[width=3in]{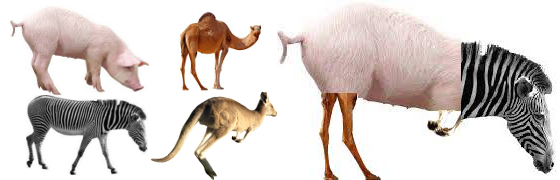}
  \caption{Example randomly selected generated photobash on the right, with the input images on the left.}
  \label{fig:generatedPhotobash}
\end{figure}

We include painting as a domain due to the long history of visual art in computational creativity, such as AARON \cite{cohen1995further} and the Painting Fool \cite{colton2012painting}. The painting problems of CrIB reflect the general problem description outlined in the introduction. 
\begin{itemize}
\item {\bf Input:} 2-6 colors as an initial knowledge base or palette.
\item {\bf Goal:} A painting that includes colors not included in the agent's initial knowledge base. The agent cannot directly access this goal painting.
\item {\bf Domain-Specific Function:} The agent is given a function \textbf{Paint} that takes as arguments two variables $x$ and $y$ (ranging from 0 to 1) that determine the location of a pixel and a color (represented in RGB format). This function then sets that pixel to the specified color on an initially blank canvas of fixed size.
\item {\bf Clear:} This function allows the agent to clear the current canvas, resetting it to an all-white image.
\item {\bf Score: } The scoring function compares the current canvas with the target painting running vector subtraction for each pixel, it then sums over these values and normalizes to a maximum of 1.0. Without an agent inventing new colors, it is impossible to score a perfect 1.0 on any of these problems. However, it is possible to get to a relatively large local maxima.
\end{itemize}

We present examples of target images in Figure \ref{fig:paintingTargets}. To generate these problems we wrote a simple combination process that takes the primary (red, green, blue), secondary, and tertiary colors of a color wheel and finds all possible additive and subtractive combinations of colors (e.g. red and blue to make purple) From there it selects either a single color combination or multiple color combinations to draw upon for each problem, creating random geometric shapes with the target colors. We sorted the final 400 questions in terms of the number of initial colors and the number of colors and shapes in the target image as a stand-in for difficulty.

\subsection{Alien Language}

We include a fictional alien language as one of our domains as a stand-in for many language domains in the field of computational creativity such as human and musical language. In addition, making use of an alien language allowed us to include one problem domain in which the answers would be less obvious to a human, given that the language would follow artificial rules without a basis in real language. This allows us to consider a human subject study as a future baseline. 

\begin{itemize}
\item {\bf Input:} A set of 3-9 words as an initial knowledge base or vocabulary.
\item {\bf Goal:} A sentence that includes words not in the initial vocabulary, representenced as a sequence of words. This sentence is not directly accessible to the agent.
\item {\bf Domain-Specific Function:} The agent is given a function \textbf{AddWord} that takes as arguments one word from the knowledge base and adds it to the current sentence. 
\item {\bf Clear:} This function allows the agent to clear the current sentence, resetting it to an empty sequence.
\item {\bf Score: } The scoring function compares the current sentence with the target sentence, giving a score of 1.0 for a perfect match, a 0.0 for no match, and a proportional score for partial matches of words in order in the sentences.
\end{itemize}

The alien language problems were generated by first generating a 2000-word vocabulary composed of randomly composing words from the characters `A', `B', `C', `D', `W', `X', `Y', and `Z' varying in length between two and twelve characters. From there we made use of arbitrary rules to compose a total of 400 target sentences varying in length between two and five words. For each target sentence we found one or two words in the sentence that could be considered combinations of between two and three other words in the vocabulary. For example ``WAZZ'' could be broken into `WA' and `ZZ'. We then sorted each problem according to the number of input words and its length as a stand-in for difficulty. For example a simple sentence might be ``WAZZ BYXBYW XDWB" with the initial knowledge base `BYXBYW', `XDWB', `WA', and `ZZ'. 

\subsection{Photobashing}

Photobashing is the practice of combining sets of input images to create a new composite image. It is a common practice for concept and key art for films, television shows, and video games. We include photobashing as it fits our general problem format and represents a real-world application of combinational creativity.

\begin{itemize}
\item {\bf Input:} A set of 2-9 images as an initial knowledge base.
\item {\bf Goal:} A goal image that represents a combination of the input images. This image or photobash is not directly accessible to the agent.
\item {\bf Domain-Specific Function:} The agent is given a function \textbf{Stamp}, which takes as arguments $x$ and $y$ variables (ranging from 0.0 to 1.0) and one of the images of the knowledge base and places this image at an $x$,$y$ location of an initially blank canvas of fixed size. Note that the agent can only add entire images from its knowledge base, meaning invention must occur to reach the goal.
\item {\bf Clear:} This function allows the agent to clear the current canvas, resetting it to a blank canvas.
\item {\bf Score: } The same as the painting scoring function.
\end{itemize}

To start generating photobashes we first gathered a palette of over eighty royalty-free stock images and photographs. We then made use of two distinct approaches to combine these images. For one we asked five human artists of a range of skill to construct photobashes. This lead to a total of 80 photobashes with a median value of 14 photobashes contributed across the five artists. An example of a human photbash can be found in Figure \ref{fig:humanPhotobash}. For the remaining 320 photobashes we constructed a simple visual grammar by breaking apart a number of images of animals into heads, torsos, front legs and back legs. We then ran a script to combine these components. We required a human to verify the coherency of each generated photobash to ensure a baseline of quality. We reran the generation process for each rejected photobash. An example of a generated photobash can be found in Figure \ref{fig:generatedPhotobash}. The problems were sorted according to number of input images used to construct the goal as a stand-in for difficulty.

\subsection{Narrative}

We include narrative as a problem domain as it represents a common area of creativity research and allows us to include a novel representation. We made use of a story or plot graph representation as it encodes the branching nature of stories \cite{weyhrauch1997guiding}. Plot graphs can be understood as a directed graph with the nodes as story events and the edges representing preconditions for events. Plot graphs represent multiple possible stories in a given domain and can generate stories by walking the graph. 

\begin{itemize}
\item {\bf Input:} A set of 2-4 distinct plot graphs
\item {\bf Goal:} A goal story represented as a sequence of events that cannot be generated from any of the input plot graphs. This story is not directly accessible to the agent.
\item {\bf Domain-Specific Function:} The agent is given a function \textbf{Submit}, which takes a single plot graph argument, and finds the closest story in the graph to the goal story. This closest story is set as the current story.
\item {\bf Clear:} This function removes any current story.
\item {\bf Score: } This function compares the current story and the target story. It returns 1.0 if the two match exactly, 0.0 if the two completely differ, and otherwise a proportional score for the number of shared events in sequence.
\end{itemize}

To begin the generation of narrative problems we first encoded ten existing published plot graphs in a common representation. We did this to ensure we did not accidentally encode too much stylistic similarity in the plot graphs. We pulled the movie, robbery, and pharmacy plot graphs from \cite{li2015learning}, the cat lover, cattle driver, stage coach and tour bus plot graphs from \cite{permar2013conceptual}, the inheritance plot graph from \cite{min2008planning}, the fantasy plot graph from \cite{mcintyre2010plot}, and the horror Anchorhead plot graph from \cite{nelson2005search}. For each plot graph we replaced the names of characters, each only had up to two, with 'A' and 'B'. We also simplified a few of the plot graphs such that each were at most 20 nodes. We then made use of amalgamation \cite{ontanon2010amalgams} to generate new plot graphs. To allow for mapping across different plot graphs we hand tagged certain event nodes with a higher-order theme (e.g. `intro', `ending', etc), additionally allowing mapping on shared words across nodes. From these plot graph amalgams we generated stories, which we then hand-checked to ensure coherency. For example a combination of fantasy and tourbus might output: ``Monster holds B captive. A slays monster. A rescues B. A departs with B. A and B get married. A and B visit a Landmark." We sorted these problems according to the number of initial plot graphs used to create the goal story's plot graph.

\subsection{Dessert Recipe}

For our final domain we chose recipes, more specifically dessert recipes, as recipes represent a common example domain for adaption and creativity. This also allowed for a second real-world domain beyond photobashing. For each dessert recipe problem the agent must invent a recipe given existing recipes.

\begin{itemize}
\item {\bf Input:} A set of 3-130 distinct recipes encoded as a recipe name and a set of ingredients (e.g. banana muffins (bananas, flour, eggs, milk, sugar)).
\item {\bf Goal:} A goal recipe distinct from all of the input recipes. This goal recipe is not directly accessible to the agent.
\item {\bf Domain-Specific Function:} The agent is given a function \textbf{Submit}, which takes a single recipe argument. This is set as the current recipe.
\item {\bf Clear:} This function removes any current recipe.
\item {\bf Score: } This function compares the current and target recipe ingredients. It returns a value between 0 and 1 dependent on the extent to which the two sets overlap.
\end{itemize}

To generate these problems we drew on the dessert dataset from \cite{vealedeja}. For each dessert we found all sets of other desserts whose ingredients could be composed to match its ingredients. From this point it was simple to randomly select a set of four hundred of these possible compositions for each problem. We then sorted these problems according to the number of initial desserts in the knowledge base as a stand-in for difficulty. This number varied massively from 3 to 142. As an example given banana muffins (bananas, flour, eggs, milk, sugar), Vanilla wafer cake (shredded coconut, flour, milk, eggs, sugar, chopped pecans, vanilla essence), and treacle tart (golden syrup, lemon zest, butter, flour) produce pound cake (butter, sugar, eggs, flour, vanilla essence).

\begin{table*}[tb]
\centering
\caption{Average output of two baselines and the random agent for each domain and across all five domains.}
\begin{tabu}{|l|c|c|c|c|c|[2pt]c|}
  \hline
   & Painting & Language & Photobash & Narrative & Dessert & Total \\
  \hline
  Null & 0.70 & 0.0 & 0.76 & 0.0 &0.0 & 0.29\\
  \hline
  Uncreative Max & 0.85 & 0.72 & 0.89  & 0.45 & 0.49 & 0.61\\
  \hline
\end{tabu}
  \label{tab:results}
\end{table*}

\begin{table*}[tb]
\centering
\caption{Scores for the presented agents and their average total.}
\begin{tabu}{|l|c|c|c|c|c|[2pt]c|}
  \hline
   & Painting & Language & Photobash & Narrative & Dessert & Total \\
  \hline
  Random Agent & -0.99 & -2.42 & -1.50 & -0.32 & -0.50 & -1.15\\
  \hline
  $GA_{100}$ & -0.99 & -1.41 & 0.02  & 0.76 & 0.35 & -0.25\\
  \hline
  $GA_{1000}$ & -0.91 & -1.19 & 0.17  & 0.81 & 0.35 & -0.14\\
  \hline
\end{tabu}
  \label{tab:results}
\end{table*}

\section{Using CrIB}

In this section we discuss how to make use of CriB. We introduce two baselines to better characterize the benchmark, introduce a scoring function that relies on one of these two baselines, and present two initial agents that attempt to solve the benchmark.

\subsection{Baselines}

In this section we demonstrate two baselines to further characterize CrIB. The baseline ``null'' represents the score of an agent that does absolutely nothing. The baseline ``Uncreative Max'' represents the best an agent could do without any invention of additional knowledge beyond the initial input knowledge base for each problem. We constructed Uncreative Max by finding the closest element of the initial knowledge base to the target concepts.

We summarize the average scores of these two baselines in Table 1. We note that the two visual domains---painting and photobashing---can achieve the highest values since they only look at pixel-by-pixel comparisons and share many white pixels. In addition, it is relatively easy to score high on the alien language domain since the goal sentences are composed mostly of words from the initial knowledge base. However, narrative and dessert generation are far less successful. Our baselines are not meant to signify any intelligence, but to provide a means for analyzing how easy it is to guess a high-scoring solution without creative reasoning if we attempt to score naively.

\subsection{Scoring Function}

The prior section demonstrates that it is possible to get high scores without creative behavior if we score naively. However, we intend this benchmark to measure a facet of creativity. Therefore we use the following scoring function for each problem domain: 

\[ Score= (NScore_{a} - NScore_{u})/(400-NScore_{u}) \]

Where $Score$ represents our final score, $NScore_a$ represents the naive score discussed for each domain above for some current agent $a$, $NScore_u$ represents the naive score discussed above for the Uncertain Max baseline. In other words an agent's actual score is the amount that it does better than Uncreative Max. We are essentially making the assumption that if the score of Uncreative Max represents uncreative computation, whatever is left must require creative computation. Because Uncreative Max makes use of all available knowledge without any invention of new knowledge, an agent may receive a negative score if it fails to make use of all of the knowledge it is initially given.

\subsection{Initial Agents}

We present two initial agents as a means of demonstrating that the problems of this benchmark are non-trivial. For the first agent we present a random agent that randomly selects a single element of the initial knowledge base and runs the domain-specific function. We note that this first agent cannot be expected to do better than Uncreative Max, but we include it in order to compare it to our second agent. Our second agent is a genetic algorithm (GA) agent, which we tested in two variations. 

The GA agent searches in the space of possibile final answers relevant to each domain (images for painting and photobashing, sentences for alien language, recipes for dessert recipe, and stories for narrative). It uses a mutation function that randomly swaps out some value of the current representation with a value from the knowledge base. It uses a crossover function that randomly selects values from two parents, selected according to current naive score, to fill in the variables of a new child representation (e.g. randomly grabbing words from two parent sentences to create a child sentence). We used a mutation rate of 0.7, and selected the 20 best parents to create 20 new children with each iteration. We created two variations on this agent based upon number of iterations and population size. For the first $GA_{100}$ we ran the GA for a maximum of 100 iterations with a population of 100 individuals. For the second $GA_{1000}$ we ran for a maximum of 1000 iterations with a population of 1000 individuals. We present the scores of all agents in Table 2. 

We note a number of interesting results comparing the scores across these agents. $GA_{1000}$ did the best, as one might expect, but did far worse than one might naively assume. The primary reason for this was that the simple mechanism by which both GA agents introduced new knowledge (random mutations and crossover) was insufficient to produce the desired combinations given the feedback of the scoring function. This is most clear in comparing $GA_{1000}$ and $GA_{100}$ in terms of the Dessert Recipes and Narrative performance. In the former there was no improvement in the score despite a tenfold increase in iterations and population. The most successful domain was Narrative, since the agent's crossover and mutation functions were well-suited to swapping out events in a story. We found with additional tests that the $GA_{1000}$ values largely represent the upper-bound of this approach, indicating that solving this benchmark is not simply a problem of longer training time.

\subsection{Ways of Using CrIB}

We include all of the discussed agents and baselines and a few additional agents on the public GitHub. Beyond reporting scores we recommend researchers make use of these given agents to draw comparisons. In particular beyond score we recommend reporting the average increase in size of the knowledge base per problem and the number of guesses or training steps necessary to achieve the reported scores. These features can allow for better comparison in terms of an agent's ability to make insightful or human-like combinations quickly. In terms of formats for reporting results we anticipate that this will depend on the agent. One clear approach would be to make use of Reinforcement Learning, which might involve reporting average score over time. Alternatively one might approach this problem with a more traditional classifier, at which point reporting training and testing error may be appropriate.

We note that one naive approach might be to hand-author knowledge for each domain. For example, simply giving an agent all primary, secondary, and tertiary colors for the painting domain. However, this goes against the spirit of the benchmark, and entirely removes any need for creative reflection or invention from an agent. 

\section{Limitations and Future Work}

We note that the benchmark at present has a number of limitations. We do not present any successful, creative agents by our own measures in this paper. The development of such agents remains the largest area of future work Further, while relatively large at first glance 2000 problems is small compared to similar benchmarks in other domains. Notably it would be trivial to expand the painting, alien language, and the dessert recipe domains to many times their current size, which one can accomplish given the GitHub generator code. However the need for human evaluation for narrative and photobashing represents a limiting factor.

There are many more possible domains we could include in this benchmark. For example music and product generation, both common computational creativity domains. We fully intend to expand CrIB in future versions.

\section{Conclusions}

We present the Creative Invention Benchmark (CrIB), a benchmark for evaluating combinational p-creativity. We demonstrate the generative process for creating the 400 problems for each of the five domains of the benchmark, and the performance of a set of baselines and agents. We make this baseline available to the general research community through GitHub, and hope that it inspires further developments in the field of computational creativity.

\section{Acknowledgments}

We gratefully acknowledge the NSF for supporting this research under NSF award 1525967. We appreciate the detailed work of the ICCC reviewers, whose insight greatly improved this final paper. In addition we would like to especially thank Jack Yardley Ingram, without whom this paper would not have been finished.

%\appendix{\LaTeX{} and Word Style Files}\label{stylefiles}

%The \LaTeX{} and Word style files are available on the ICCC-13
%website, {\tt http://computationalcreativity.net/iccc2013/}.
%These style files implement the formatting instructions in this
%document.

%The \LaTeX{} files are {\tt iccc.sty} and {\tt iccc.tex}, and
%the Bib\TeX{} files are {\tt iccc.bst} and {\tt iccc.bib}. The
%\LaTeX{} style file is for version 2e of \LaTeX{}, and the Bib\TeX{}
%style file is for version 0.99c of Bib\TeX{} ({\em not} version
%0.98i).

%The Microsoft Word style file consists of a single template file, {\tt
%iccc.dot}. 

%These Microsoft Word and \LaTeX{} files contain the source of the
%present document and may serve as a formatting sample.  

\bibliographystyle{iccc}
\bibliography{iccc}

\end{document}